\documentclass{article}

     \usepackage[final,nonatbib]{neurips_edu_2020}

\usepackage[utf8]{inputenc} 
\usepackage[T1]{fontenc}    
\usepackage{hyperref}       
\usepackage{url}            
\usepackage{booktabs}       
\usepackage{amsfonts}       
\usepackage{nicefrac}       
\usepackage{microtype}      
\usepackage{graphicx}

\title{Explainable Knowledge Tracing Models for Big Data: Is Ensembling an Answer?}

\author{%
    Tirth Shah\\
    Playpower Labs\\
    Gujarat, India \\
    \texttt{tirth.shah@playpowerlabs.com} \\
    \And
    Lukas Olson \\
    Infinite Campus \\
    Minnesota, United States \\
    \texttt{lukas.olson@infinitecampus.com} \\
    \And
    Aditya Sharma \\
    Playpower Labs \\
    Gujarat, India \\
    \texttt{aditya@playpowerlabs.com}
    \And
    Nirmal Patel \\
    Playpower Labs \\
    Gujarat, India \\
    \texttt{nirmal@playpowerlabs.com} \\
}

\begin{document}

\maketitle

\begin{abstract}
In this paper, we describe our Knowledge Tracing model for the 2020 NeurIPS Education Challenge. We used a combination of 22 models to predict whether the students will answer a given question correctly or not. Our combination of different approaches allowed us to get an accuracy higher than any of the individual models, and the variation of our model types gave our solution better explainability, more alignment with learning science theories, and high predictive power.
\end{abstract}

\section{Introduction}

The widespread presence of affordable digital learning platforms is making quality education available to a large number of students. Many students can now freely access open educational content that is aligned with their school curriculum. Online learning platforms contain many different types of resources for students: instructional videos, reading materials, interactive explore tools, formative and summative assessments, etc. Most of the online learning systems extensively track how students use these resources and advance their learning over time. The scale of online education is now ever increasing, and it is providing researchers with massive datasets that can be used to make learning more efficient and engaging and advance the science of learning.

We can use big educational data to empower adaptive learning algorithms. These algorithms can deliver students educational resources they need the most. Personalized learning is one of the core promises of the education technology and it gives us an opportunity to give each student the help that they need. There are many different personalized learning algorithms, and a vast number of them have evolved from the ideas around the theory of Knowledge Components and \cite{kcwebpage} Knowledge Tracing \cite{corbett1994knowledge}. One of the key goals of the Knowledge Tracing (KT) algorithms is to predict whether a student will complete an unseen educational task correctly or not. We can also use IRT algorithms to predict student responses on unseen items \cite{hambleton1991fundamentals}. IRT algorithms also offer other metrics that can be used to assess question quality.

Algorithms to predict student response on an unseen question fall across a spectrum with two poles: Theoretically Grounded and Statistically Derived. The former pole of algorithms comes from our knowledge of human cognition. Bayesian Knowledge Tracing and Deep Knowledge Tracing \cite{piech2015deep} are two examples of the more Theoretically Grounded algorithms. They are derived respectively from the ACT-R theory of cognition and the connectionist model of brain. Statistically Derived algorithms do not have any specific relation to how cognition works and how students learn. Latent variable based algorithms such as IRT and HMMs can fall somewhere in between the two extremes. There is no general consensus in the Educational Data Mining community about which methods are best \cite{khajah2016deep,gervet2020deep}. Our work in this paper combined models of different types to build an ensemble. We believe that ensemble models that combine different approaches are likely to be more robust in the real environment.

\section{NeurIPS 2020 Education Challenge}

In July 2020, Eedi \footnote{https://eedi.com} launched a predictive modeling challenge in the NeurIPS conference focused on the multiple-choice diagnostic questions \cite{wang2020diagnostic}. A diagnostic MCQ is a question with four options, where exactly one option is correct and three options are incorrect. The incorrect options highlight specific misconceptions related to the skill that is being assessed in the question. The competition had four distinct tasks: \textbf{Task 1} was a Knowledge Tracing task where the goal was to predict whether a student will answer a given question correctly or not, \textbf{Task 2} was aimed at predicting which of the four MCQ options the student will select, \textbf{Task 3} was around predicting the quality of a given question, and \textbf{Task 4} required building an intelligent tutor like interactive predictive model. Given the size of the dataset and complexity of the problem, we almost exclusively focused on Task 1. The rest of the paper describes the data and our solution to the Task 1.

\section{Data}

\begin{figure}[t]
\centering
\includegraphics[width=13cm]{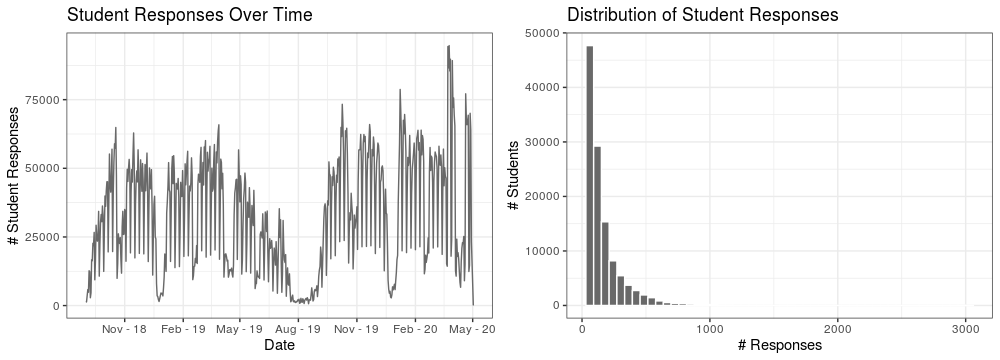}
\caption{Left: Usage of the learning platform over time, this is a very typical online education heartbeat. You can see an increased peak towards the right of the plot during the COVID crisis. Right: A distribution of the \# of of responses per student. This is also a very typical ski-slope distribution that shows that some educational resources get used a lot while others remain in oblivion.}
\end{figure}

The competition organizers provided the data consisting of student responses to multiple-choice diagnostic questions. For task 1, there were a total of 15.8 million student responses in the training dataset and 3.96 million student responses in the test dataset. The primary data table had information about unique identifiers for students, questions, and the student responses, a binary variable for whether the student response was correct or not, the correct answer option to the question, and the student's answer option to the question. More description of the data is given in the table below:
\vspace{0.5cm}
\begin{center}
\begin{tabular}{ |c|c|c|c|c|c| } 
 \hline
 \# Users & \# Questions & \# Responses & \# Skills & \# Groups & \# Quizzes \\
 \hline
 118971 & 27613 & 19834814 & 388 & 11844 & 17305 \\
 \hline
\end{tabular}
\end{center}
\vspace{0.5cm}
The competition organizers also provided secondary metadata related to users, questions and each of the answers. Student metadata had gender, date of birth and the information about whether the student was receiving a free and reduced lunch at the school or not. Several studies have shown that free and reduced lunch indicator is negatively correlated with student outcomes \cite{lomas2020}. Question metadata provided a mapping between questions and skills. The skills has multiple levels and a tree-like hierarchy. Answer metadata had information about each of the answer. They contained the timestamp for each response, an answer confidence score (this column was nearly empty), a unique identifier of the group to which student belonged, a unique identifier of the quiz to which the question belonged, and a unique identifier for scheme of work for the question.

\section{Models for Task 1}

\subsection{Feature Engineered Models}
In this subset of models, we relied on creating features and training different types of predictive modeling algorithms on those features. For each of the student response, we extracted a total of 14 features. We ensured that for every student response, only the data before that response was considered, since future student interactions cannot be used to predict the past. The features were based on the learning level of the student at each skill hierarchy, difficulty of the question that the student is answering, overall difficulty of the quiz, learning level of the peers of the student, a momentum in student learning process which was calculated by taking a decayed average of most recent responses, time passed since the student last attempted a question of the same skill, etc. This powerful set of features helped us achieve better prediction accuracy. With the help of our features, we trained a total of 18 statistical models.

For the first 16 models in this subset, we used 4 different classes of algorithms combined with 4 different model configurations. Algorithms used were Gradient Boosted Trees (we used the Catboost package \footnote{https://catboost.ai}), KNN, Naive Bayes Classifier and Bayesian Generalized Linear Model. Different model configurations were designed to compute the models at different level of granularity. The four levels used were question, user, group, and quiz. The other 2 models were fit on the full training dataset and algorithms used for them were Catboost and KNN. The idea was to use a diverse set of statistical methods and different modelling frameworks which can capture different information from the training data so as to obtain a set of uncorrelated prediction probabilities which can then be combined to give us a better prediction accuracy. 

\subsection{Deep Knowledge Tracing Based Model}
Given the success of attention-based models on knowledge tracing problems \cite{pandey2019self,ghosh2020context}, we included an encoder-only model to impute missing answers—an approach similar to the masked language model BERT \cite{devlin2018bert}. Each example sequence consisted of all questions attempted by a user and their corresponding answer. We trained two embeddings, one for each question and one for each answer value (correct, incorrect, masking, and padding), concatenated them, and added a positional encoding to each time step. We fed the corresponding sequence into a stack of 6 encoder blocks which transformed the sequence to predict the target answer at each time step. Similar to Devlin et. al.’s approach in \cite{devlin2018bert}, we randomly masked 20\% of the time steps; however, rather than masking the entire time step, we replaced only the answer token with a masking token since the information about which question was attempted is crucial in predicting proficiency. Each encoder block contained 4 attention heads and a pointwise feed forward network with an approximated GELU activation \cite{hendrycks2016bridging}. We used a latent vector dimensionality of 512, batch size of 64, and maximum sequence length of 400.

\subsection{IRT and Knowledge Component Theory Inspired Models}
This subset of the models included three models. The first model modeled student knowledge as a skill profile considering skills as Knowledge Components \cite{kcwebpage}. We generated a matrix of students as rows and skills as columns, where the cell value was student average score in the given skill. The matrix was incomplete, so we used collaborative filtering algorithm to impute the missing values. The imputed information about student skill proficiency was then used to predict the student response on a given question. The second model used 2-PL IRT model \cite{hambleton1991fundamentals} to predict how students would respond to given questions. Using IRT in an extremely sparse dataset is challenging, so we ran the IRT algorithm quiz by quiz. The third model first generated new quiz IDs by using question clustering. The questions were clustered using hierarchical clustering. We provided an inter-question distance matrix to the HC algorithm where distance between the questions was defined as the negative of how many students took that question in common. The new quiz IDs led to less sparse item response matrices, on which we then used the used the same method as the second model.



\subsection{Model Ensembling}
Ensembling turned out to be a powerful approach for improving accuracy. We tried submitting all of the above methods separately, but none of them could beat the accuracy estimates of the final ensemble model. Our best solution was a weighted probabilistic ensemble of 22 models. 

\section{Results}
Our final ensemble model gave us an accuracy of 76.17\% on the public test set of Task 1 and 76.21\% on the private test set of Task 1. Rest of the results described in this section are on the public test set. One of our initial models which included all of the features and was trained on all of the data using Catboost gave us an accuracy of 75.99\%. A Catboost Model on the full training data which comprised of 8 features gave us an accuracy of 73.85\%. For the same training data of 8 features, a Random Forest model gave us an accuracy of 72.52\% and a Logistic Regression model gave us an accuracy of 73.53\%. An ensemble of the models described in the section 4.3 gave us an accuracy of 73.64\%. We trained our models using both CPU and GPU. Catboost supports GPU training which came in really handy to quickly experiment with different feature based models. Some Catboost models took 10-15 mins to train, while others took hours. Random Forest algorithm took the most amount of training time which was nearly 15 hours.

\section{Discussion}
We discovered that taking an ensembling approach to Knowledge Tracing allowed us to use high accuracy complex models such as Neural Networks along with more explainable models where we were able to look at importance of different features and take an approach that was in line with learning science research. It is nearly impossible to do well with just one approach, and we hope that our case study helps other Educational Data Mining teams take a more hybrid approach that both works better and has parts that people can understand and make sense of.

\begin{ack}

We would like to thank Playpower Labs and Infinite Campus for their interest in education research and support of our involvement in this competition. We also acknowledge the competition organizers for hosting the challenge and providing this valuable data set to the community.

\end{ack}

\small
\bibliographystyle{IEEEtran}
\bibliography{main}

\end{document}